\documentclass{article} 
\usepackage{iclr2027_conference,times}

\usepackage{amsmath,amsfonts,bm}

\newcommand{\method}[0]{ORMA}

\newcommand{\pose}{\theta}

\newcommand{\trasl}{\textbf{t}}
\newcommand{\grot}{R_g}

\newcommand{\image}[1]{F_{#1}}

\def\eqref#1{equation~\ref{#1}}

\def\1{\bm{1}}

\DeclareMathAlphabet{\mathsfit}{\encodingdefault}{\sfdefault}{m}{sl}
\SetMathAlphabet{\mathsfit}{bold}{\encodingdefault}{\sfdefault}{bx}{n}

\usepackage{amsmath}
\usepackage{amssymb}
\usepackage{fix-cm}
\usepackage{mathtools}
\usepackage{bm}
\usepackage{hyperref}
\usepackage[capitalise,nameinlink,noabbrev]{cleveref}
\usepackage{url}
\usepackage{graphicx}
\usepackage{booktabs}
\usepackage{multirow}
\usepackage{url}
\usepackage{svg}
\usepackage{adjustbox}
\usepackage{makecell}
\usepackage{booktabs}
\usepackage{xcolor}
\usepackage{float}
\usepackage[accsupp]{axessibility}  %
\usepackage{caption}

\usepackage[percent]{overpic}
\usepackage{graphicx}
\usepackage{tikz}

\title{ORMA: Optimization-based Monocular \\ 4D Reconstruction of Articulated Animals}

\author{%
\small
Xuyi Hu$^{1^{*}}$ \hspace{0.4em}
Francesco Palandra$^{2^{*}}$ \hspace{0.4em}
Shangzhe Wu$^{1}$ \hspace{0.4em}
Daniel Cremers$^{3,4}$ \hspace{0.4em}
Riccardo Marin$^{3,4}$ \hspace{0.4em}
Silvia Zuffi$^{2}$
\\[0.4em]
\footnotesize
$^{1}$University of Cambridge \qquad
$^{2}$IMATI-CNR, Milan, Italy \qquad
$^{3}$Technical University of Munich, Germany \\
\footnotesize
$^{4}$Munich Center for Machine Learning, Germany
}

\iclrfinalcopy 
\begin{document}

\maketitle

\begingroup
\renewcommand{\thefootnote}{\fnsymbol{footnote}}
\footnotetext[1]{Equal contribution.}
\endgroup

\begin{center}
    \includegraphics[width=\textwidth]{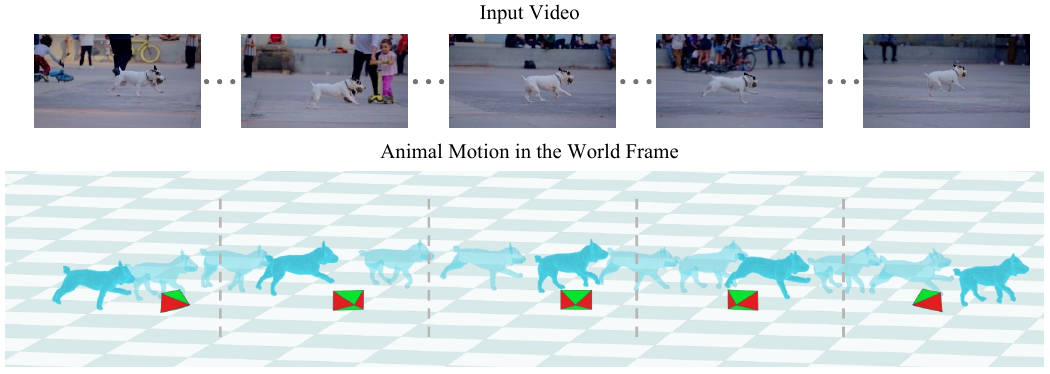}
    \captionof{figure}{4D Reconstruction of Articulated Animals from Videos in the Wild. We present ORMA, an optimization-based framework that reconstructs the shape and motion of articulated animals in a shared world coordinate frame from monocular videos. 
    \label{fig:teaser}}
\end{center}

\label{sec:abs}
\begin{abstract}
Recovering articulated 4D representations of animals from monocular videos remains challenging due to the large diversity of quadruped morphologies and lack of animal 4D supervision data.
Existing learning-based reconstruction methods operate on individual images and rely on synthetic or model-fitted 3D supervision, which inherits the constraints of strong parametric priors and limits generalization to out-of-distribution species. When applied to out-of-distribution animals, they often recover a plausible pose while producing inaccurate geometry because the underlying shape model cannot faithfully represent the observed instance. We present \method, a training-free reconstruction framework that decouples articulation from shape, using the predicted pose as reference for optimization while leveraging generative 3D priors for accurate shape reconstruction.
Given a reference image, we reconstruct the animal geometry and register it to the parametric model SMAL{+}, yielding an articulated shape adapted to the observed instance. We then combine per-frame articulated pose estimates with globally consistent camera poses to recover animal motion in a shared world coordinate frame, and further refine the reconstruction using self-supervised DINO correspondences and temporal consistency. To enable quantitative evaluation, we introduce PAW4D, a synthetic multi-species benchmark with ground-truth 3D geometry and camera motion. Experiments on PAW4D, PFERD, and challenging in-the-wild videos demonstrate that \method{} improves reconstruction accuracy while recovering globally consistend animal motion 
across diverse quadruped species.


\end{abstract}

\section{Introduction}
\label{sec:intro}

Accurate 4D reconstruction of animals from monocular video, encompassing both 3D shape and motion estimation, remains a central challenge in computer vision with important implications for ecology, animation, biomechanics, and robotics. 
Parametric models able to represent the articulated shape of animals offer a structured foundation for 4D reconstruction, where learned shape spaces and motion priors enable temporally consistent estimation of an animal’s geometry and movement from ambiguous monocular data.
Importantly, these models encode articulated motion in a biomechanical representation, typically as 3D joint rotation angles, facilitating motion analysis and retargeting.

Unlike humans, whose body shape and motion have been extensively modeled through large-scale datasets and parametric representations~\citep{SMPL:2015,SMPL-X:2019}, which have supported the recovery of global human motion~\citep{ye2023decoupling}, animals are difficult to capture, and exhibit far greater morphological diversity and motion complexity. 
Human-based, controlled, large-scale capture setups generate accurate supervision datasets but this approach is infeasible to replicate for animals, although similar strategies have been applied to specific species, such as horses~\citep{Zuffi_2024_CVPR, Li2024PFERD}. 
The lack of large-scale 3D and motion data makes it difficult to build statistical models that generalize across species while capturing both fine-grained deformations and natural kinematics. Consequently, model-based methods, such as those relying on parametric models learned from small-scale training sets, like SMAL~\citep{Zuffi:CVPR:2017}, struggle to capture unseen species or out-of-distribution poses. 
Recent multi-species learning-based 3D reconstruction methods are often trained on synthetic animal datasets with 3D supervision, leveraging strong parametric priors to recover articulated motion for in-distribution species~\citep{lyu2025animer,niewiadomski2025ICCVgenzoo}. However, when applied to out-of-distribution animals, they often recover a plausible pose while producing inaccurate geometry because the
underlying shape model cannot faithfully represent the observed instance.
Model-free methods that infer shape and motion directly from image evidence without assuming a pre-defined template offer greater flexibility, but often produce temporally inconsistent or physically implausible results due to the absence of explicit 3D priors to support the reconstruction of unseen parts~\citep{BANMO}. 
Bridging this gap and combining the interpretability and stability of model-based approaches with the adaptability of model-free ones remains an open problem. Recent progress in 3D generative models offers a promising alternative: geometric priors learned from large-scale 3D assets~\citep{zhao2025hunyuan3d20} can recover plausible instance-specific shapes from a reference frame. Registering these shapes to a rigged quadruped model then yields animatable representations compatible with existing pose estimators.
Moreover, current methods for model-based 4D reconstruction of animals mostly recover pose in camera frame, often assuming a weak perspective camera. For quadrupeds, whose bodies often extend along the camera axis, this can lead to inaccurate shape and pose estimates. Recent advances in camera trajectory and intrinsics estimation provide an opportunity to recover global motion even from monocular in-the-wild video.

We present \method{}, a hybrid reconstruction method for 4D reconstruction that leverages the advantages of image-to-3D generative priors while retaining the structured pose representation of the SMAL+ model.
In \method{}, we aim at recovering animal motion in world frame with the true global camera.
To achieve this, we formulate 4D reconstruction as an unsupervised optimization problem over time. 
Given a video, we estimate the 3D animal shape from a reference frame with an image-to-3D generator~\citep{zhao2025hunyuan3d20}. We estimate the camera with a recent approach for camera estimation from video sequences~\citep{li2025megasam} and adapt the generated shape such that its projection matches the image under the estimated camera. We express the 3D shape as a SMAL+ model~\citep{zuffi2024awol} instance through registration, and initialize the pose of the obtained model in each frame with a quadruped pose estimator~\citep{lyu2025animer}, providing a strong starting point for optimization. 
We then employ a set of visual losses to better align the model to the data, employing a novel dense feature correspondences loss to establish semantically meaningful alignments between 3D surface points and image pixels. 

Methods for 4D animal reconstruction have been evaluated so far on 2D test data due to the lack of video datasets with accurate 3D ground truth. While PFERD~\citep{Li2024PFERD} provides real horse videos with diverse shapes and accurate motion-capture-based 3D ground truth, it is limited to horses. To make progress, we introduce the first multi-species synthetic dataset PAW4D that includes a variety of species performing short actions.
Through experiments 
on our synthetic benchmark and on real videos with 3D ground truth~\citep{Li2024PFERD}, we show that \method{} recovers temporally coherent articulated geometry and motion in a consistent global frame from monocular video.
Applied at scale to in-the-wild videos, the framework could support the construction of large and diverse 3D animal-motion datasets, analogous to recent efforts for humans~\citep{BEDLAM2}. 
Such datasets 
could be used to learn animal motion models and synthesize realistic training data for future 3D and 4D reconstruction methods.
Ultimately, our framework bridges model-based and model-free paradigms, retaining the structural advantages of parametric models while adapting flexibly to diverse real-world animal forms and motions.

In summary, ORMA addresses the scarcity of 4D animal data through a modular optimization framework that composes complementary pretrained priors without requiring task- or species-specific training. Specifically, we introduce a shape--articulation decoupling formulation that combines instance-specific geometry with structured SMAL{+} kinematics, a Robust Semantic Correspondence (RSC) module for reliable image-to-surface alignment, and a unified optimization objective for world-space geometry and motion recovery. We further introduce PAW4D, a synthetic multi-species benchmark with ground-truth geometry and camera motion for quantitative evaluation.

%

\section{Related Work}
\label{sec:related}


\noindent\textbf{3D Animal Reconstruction from Images.}
The 3D reconstruction of animals has evolved along two primary paradigms: model-free and model-based approaches.
Model-free methods aim to recover 3D geometry with minimal structural body assumptions. Given the significant inter-species variation in morphology, such flexibility is appealing. Early work, such as CMR~\citep{cmrKanazawa18}, reconstructed birds by deforming a spherical template. Subsequent approaches, including LASSIE~\citep{Yao2022lassie}, MagicPony~\citep{Wu2023magicpony}, and 3D-Fauna~\citep{Li20243Dfauna}, learned articulated 3D representations from image collections. 
While model-free methods offer flexibility and category-level generalization, they typically lack explicit semantic control over skeletal pose, which limits their suitability for structured behavioral analysis or biomechanical interpretation.
Model-based approaches instead assume access to a predefined 3D template or a model, whose shape and pose parameters are estimated from images or video. This paradigm provides representations that are particularly valuable for downstream tasks such as conformation analysis, health assessment, and motion tracking.
A major milestone was the introduction of SMAL~\citep{Zuffi:CVPR:2017}, a multi-animal model learned from toy scans that captures articulated shape variation across quadrupeds. SMAL is widely adopted~\citep{Zuffi_2019_ICCV}~\citep{biggs_2019, biggs2020left, rueegg2022barc}. Species-specific parametric models have been developed for dogs and horses~\citep{ruegg2023bite,li2021hsmaldetailedhorseshape}, while related model-based approaches have also been extended to birds~\citep{badger_2020_eccv,Wang_2021_CVPR} and dolphins~\citep{baieri2025modelbasedmetric3dshape}.
Recent efforts in creating parametric animal models include the horse model learned from real 4D scans introduced in~\citep{Zuffi_2024_CVPR} and the SMAL+ introduced in AWOL~\citep{zuffi2024awol}, enhancing the original shape space with additional 3D scan data. While early works focused on pose and shape estimation for single species, the most recent efforts~\citep{lyu2025animer,niewiadomski2025ICCVgenzoo,yu2026prima} are multi-species regression networks trained with synthetic 3D data and eventually 2D supervision. SAM3D Animal~\citep{hu2026sam} further extends this paradigm to promptable multi-species and multi-instance reconstruction.


\noindent\textbf{4D Animal Reconstruction from Videos.}
Animal 4D reconstruction has been explored through both model-free and model-based approaches. Among model-free methods, ViSER~\citep{yang2021viser}, LASR~\citep{yang2021lasr}, BANMo~\citep{BANMO}, DOVE~\citep{wu2023dove}, and PPR~\citep{yang2023ppr} jointly optimize geometry, articulation, and appearance from monocular videos, often using neural implicit or part-based representations to model non-rigid deformation. More recently, Ponymation~\citep{sun2024ponymation} and 4D-Fauna~\citep{zhao2025webscaleanimal4dfauna} extend
3D-Fauna~\citep{Li20243Dfauna} to videos and reconstructs articulated animal shape and motion from large-scale in-the-wild sequences. Model-based approaches instead exploit structured parametric animal representations. RAC~\citep{yang2023rac} disentangles instance shape and temporal motion from monocular videos, and PADR~\citep{LiaoPADR} reconstructs deformable objects using image-to-3D priors and deformable 3D Gaussians. 
AnimalAvatar~\citep{AnimalAvatars2024} and 4DAnimal~\citep{zhong2026_4danimal} reconstruct dogs from monocular videos using optimization-based formulations with semantic correspondence cues. AniMer+~\citep{lyu2025animerunifiedposeshape+} extends image-based parametric reconstruction to video, while 4DEquine~\citep{lyu20264dequine} focuses on horses by disentangling 4D reconstruction into temporally coherent motion estimation and static appearance reconstruction. WildAni4D~\citep{cho2026wildani4d} further introduces a video-native framework that predicts temporally coherent animal meshes and global trajectories in the world frame. Kirin~\citep{zhao2026kirin} further reconstructs large-scale 3D quadruped motion from in-the-wild videos to build motion priors for downstream animal motion generation and animation. 

\noindent\textbf{Datasets for 3D and 4D Animal Reconstruction.}
Early animal datasets were primarily developed for 2D animal pose estimation and typically provide keypoint annotations for animals. Existing animal datasets vary considerably in taxonomic coverage: many are species-specific, which target individual categories such as dogs~\citep{biggs2020left}, horses~\citep{mathis2021pretraining}, Macaque~\citep{labuguen2021macaquepose}, birds~\citep{wah2011caltech}, pigs~\citep{an2023three} or tigers~\citep{li2019atrw}, while others aim to cover a broader range of animal species~\citep{aamir2026wilddepth,cao2019cross,yu2021ap,banik2021novel,yang2022apt,ng2022animal}. However, these datasets primarily provide image-space supervision and therefore cannot directly support or evaluate 3D shape and motion reconstruction. Several datasets further introduce 3D supervision through multi-view capture~\citep{joska2021acinoset}, motion capture~\citep{kearney2020rgbd}, or model fitting~\citep{xu2023animal3d}. To alleviate the cost of acquiring real 3D annotations, synthetic datasets can provide an alternative source of scalable 3D supervision~\citep{niewiadomski2025ICCVgenzoo,choi2026everydog,shooter2024digidogs,hu2026sam}. Beyond static 3D supervision, recent datasets increasingly capture temporal animal motion. PFERD~\citep{Li2024PFERD}, DogMo~\citep{wang2025dogmo}, and InterPet4D~\citep{peng2026interpet4d} provide controlled multi-view or motion-capture recordings of horses and dogs, while CoP3D~\citep{sinha2023common} and AiM~\citep{zhao2025webscaleanimal4dfauna} collect in-the-wild videos for dynamic animal reconstruction. Redirect4D-Bench~\citep{cao2026redirect4dbench} shows animal videos with pseudo-4D geometry and camera trajectories for dynamic videos. Synthetic datasets, including DeformingThings4D~\citep{li20214dcomplete}, VarenPoser~\citep{lyu20264dequine}, and WildAni4D-Gen~\citep{cho2026wildani4d}, further provide scalable dense supervision for animal geometry, motion, and camera dynamics.




\section{Methods}
\label{sec:method}
\subsection{Preliminary}
\noindent\textbf{SMAL+.}
\label{sec:SMAL}
We use SMAL{+}~\citep{zuffi2024awol}, an extended version of the SMAL model~\citep{Zuffi:CVPR:2017}, as our articulated animal representation. Following the formulation of SMPL~\citep{SMPL:2015}, SMAL{+} represents animal shape and pose using a deformable mesh model. It retains the same articulated formulation as SMAL, but learns a substantially richer shape space from an expanded set of 145 registered 3D animal scans. The model is defined by a triangular template mesh $\mathbf{v}_t$ with $n_V$ vertices,
a linear shape space represented by a matrix
$\mathbf{B} \in \mathbb{R}^{3n_V \times n_B}$ containing $n_B$ shape basis vectors,
a joint regressor $\mathbf{J}_r$ that maps mesh vertices to a set of $n_J$ skeletal joint locations,
and a skinning weight matrix $\mathbf{W}$.
In SMAL{+}, we use $n_B=145$, such that the shape parameters are
$\boldsymbol{\beta} \in \mathbb{R}^{145}$.

\subsection{ORMA} 


To reconstruct animal motion in a consistent world coordinate frame, our framework combines coherent camera estimation, explicit geometric supervision, and robust semantic correspondence. We use MegaSaM~\citep{li2025megasam} to estimate camera poses consistently across video, together with per-frame depth maps that provide explicit 3D geometric supervision. To further guide articulated pose optimization, we establish semantic correspondences using DINO features~\citep{simeoni2025dinov3}. Rather than relying on dense correspondences, we retain only sparse, high-confidence matches, which provide robust semantic cues for aligning the reconstructed animal with the observations.


We denote a SMAL{+} model instance by
$\mathcal{M}=S(\boldsymbol{\beta},\boldsymbol{\theta},
\mathbf{t},\mathbf{R}_g)=(\mathbf{V},\mathbf{F})$,
where $\mathcal{M}$ is a triangular mesh with vertices
$\mathbf{V}\in\mathbb{R}^{3889\times3}$ and faces
$\mathbf{F}\in\mathbb{N}^{7774\times3}$.
The shape parameters are $\boldsymbol{\beta}\in\mathbb{R}^{145}$, while
$\boldsymbol{\theta}\in\mathbb{R}^{3(J-1)}$ denotes the articulated pose
in axis-angle representation, with $J=35$ body joints.
The global rigid transformation is parameterized by
$\mathbf{t}\in\mathbb{R}^{3}$ and $\mathbf{R}_g\in SO(3)$. 


For brevity, we indicate a model instance as $S$ and omit its parametrization when it is clear from context.
The input to our method is a video
$\mathbf{F} \in \mathbb{R}^{H \times W \times 3 \times N}$
composed of $N$ frames
$\mathbf{F}_i \in \mathbb{R}^{H \times W \times 3}$,
each of height $H$ and width $W$.
We denote the binary foreground mask of the animal in frame $\mathbf{F}_i$,
obtained using SAM~\citep{kirillov2023segment}, as
$M(\mathbf{F}_i) \in \{0,1\}^{H \times W}$.
We define a differentiable image projector as:
\begin{equation}
    \pi(\,\cdot \mid C) :
    \mathcal{M} \mapsto I_C^{\mathcal{M}}
    \in \mathbb{R}^{H \times W},
\end{equation}
which renders a triangular mesh $\mathcal{M}$ under camera $C$.
When the camera is clear from context, we simply write
$\pi(\,\cdot\,)$.

{
\setlength{\intextsep}{7pt}

\begin{figure}[H]
    \centering
    \includegraphics[width=1.0\linewidth]{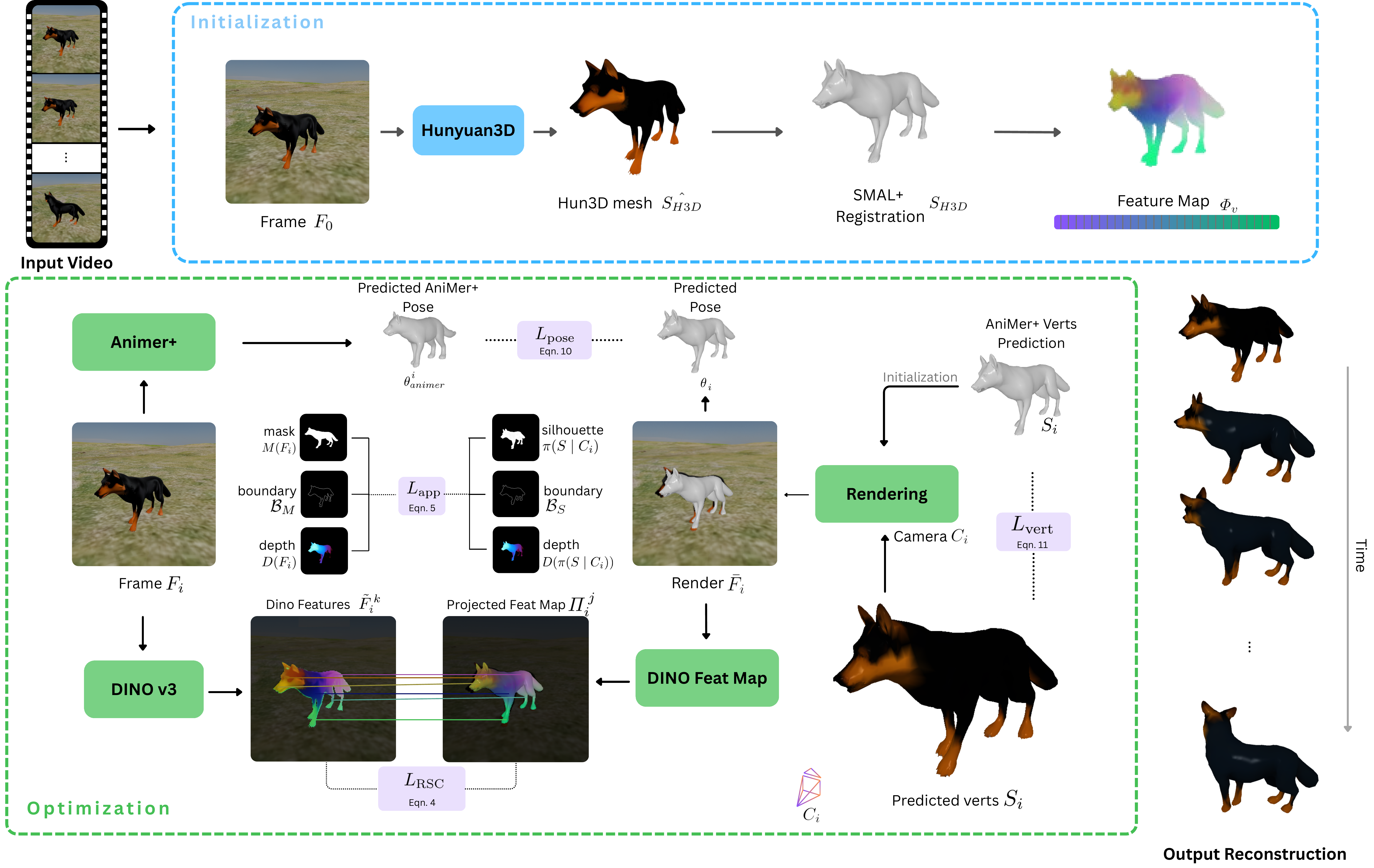}
    \caption{ORMA takes a video as input and produces a 4D reconstruction of the animal, with a trajectory in world coordinates (bottom right). As initialization, ORMA leverages Hunyuan3D~\citep{zhao2025hunyuan3d20} to obtain a reliable shape estimation of the first frame (blue box). After registering SMAL+~\citep{zuffi2024awol}, we perform a frame-by-frame optimization (green box). Unlike previous works, ORMA relies on MegaSaM~\citep{li2025megasam} to robustly estimate camera parameters across the entire video, along with a depth map. Additionally, we design a Robust Semantic Correspondence module using DINO~\citep{simeoni2025dinov3} to produce a robust set of sparse correspondences between the frames and the SMAL+ registration. Equations are provided in the appendix.
    }
    \label{fig:architecture}
\end{figure}
}

\noindent\textbf{Initialisation.}
We initialize each frame with AniMer{+}~\citep{lyu2025animerunifiedposeshape+} estimates of animal shape, pose, and camera, then use MegaSaM~\citep{li2025megasam} for video-wide camera poses and SAM~\citep{kirillov2023segment} for per-frame animal masks.

\noindent\textbf{Accurate 3D Shape Reconstruction.}
Our goal is to obtain an accurate 3D shape of the target animal. AniMer{+} predicts shape in the SMAL parameter space, which may not adequately capture the morphology of a specific species, breed, or individual. We therefore reconstruct a textured 3D mesh from the first frame where the animal is fully visible and the Animer{+} estimate is accurate using Hunyuan3D~\citep{zhao2025hunyuan3d20}.
We rigidly align this reconstruction with the posed AniMer{+} mesh using multi-start ICP, avoiding poor local minima (e.g., aligning the head with the tail). We first adjust the scale of the Hunyuan3D reconstruction using the estimated depth and camera parameters, such that its 3D mesh is consistent with the
observed animal in the scene. We then fit SMAL+ to the aligned and scaled reconstruction by optimizing the
shape coefficients $\boldsymbol{\beta}$, articulated pose $\boldsymbol{\theta}$, global rotation $\mathbf{R}_g$, and translation $\mathbf{t}$ under a Chamfer objective. This yields fitted shape parameters $\boldsymbol{\beta}_{H3D}$. Starting from this parametric fit, we further refine the geometry directly at
the vertex level using Chamfer optimization with ARAP regularization~\citep{sorkine2007rigid}, allowing local shape adaptation beyond the SMAL+ shape space. The fitted shape parameters $\boldsymbol{\beta}_{H3D}$ and the resulting vertex-level shape refinement are kept fixed throughout the sequence, while the articulated pose, global rotation, and world-space translation are optimized for each frame. For each frame, we then recover the articulated pose $\boldsymbol{\theta}$, global rotation $\mathbf{R}_g$, and world-space translation $\mathbf{t}$, initialized from the per-frame AniMer{+} pose and the fixed shape \(S_{H3D}\).

\subsection{Loss Functions}
\label{sec:loss}

\noindent\textbf{Total Loss.}
With the instance-specific shape $\boldsymbol{\beta}_{H3D}$ fixed, we optimize the articulated
pose $\pose$, global rotation $\grot$, and world-space translation $\trasl$.
Our optimization combines image-space geometric alignment, semantic
correspondence, depth supervision, and motion regularization:
\begin{equation}
\label{eq:total}
\begin{split}
L_{\mathrm{total}}
={}&
L_{\mathrm{app}}
+w_D L_{\mathrm{RSC}}
+w_p L_{\mathrm{pose}} 
+
w_d L_{\mathrm{depth}}
+\lambda_{\mathrm{temp}} L_{\mathrm{temp}},
\end{split}
\end{equation}
where $w_\star$ denotes the corresponding loss weights.
$L_{\mathrm{app}}$ aligns the rendered silhouette and boundaries with the observed
foreground mask, while $L_{\mathrm{RSC}}$ provides semantic correspondence between
the articulated surface and the image.
$L_{\mathrm{pose}}$ regularizes the solution toward the AniMer{+} pose initialization,
and $L_{\mathrm{depth}}$ constrains both surface geometry and global 3D placement using
MegaSaM depth estimates. Finally, $L_{\mathrm{temp}}$ encourages temporally smooth translation, rotation, and motion.

\subsection{Robust Semantic Correspondence}
\label{sec:RSC}

Silhouette and depth supervision constrain geometric alignment but do not
explicitly establish semantic correspondence between body parts.
Consequently, configurations with similar projections but incorrect limb
assignments may remain local minima.
Previous animal reconstruction methods use explicit surface correspondences,
such as Continuous Surface Embeddings (CSE)~\citep{AnimalAvatars2024,rueegg2022barc},
but we observe that these correspondences can be unstable across frames,
particularly around articulated extremities.
We instead use DINOv3~\citep{simeoni2025dinov3} features to construct sparse, high-confidence local
correspondences between the reconstructed animal and each video frame.

\noindent\textbf{Feature baking.}
We render the aligned Hunyuan3D reconstruction from $V$ viewpoints and extract
DINOv3 feature maps, which are upsampled to the image resolution using
AnyUp~\citep{wimmer2025anyup}.
The features from all visible views are back-projected and averaged on the
Hunyuan3D mesh vertices.
We then transfer these descriptors to the fitted SMAL{+} topology using
nearest-neighbour association in 3D.
We denote the resulting feature-decorated template as
$\mathcal{F}_{H3D}=\{\mathbf{f}_u\}$.

\noindent\textbf{Per-frame robust semantic correspondences.}
For each frame $i$, we project the visible vertices of $S_{H3D}$, each
carrying its baked descriptor $\mathbf{f}_u$, onto the image.
For each projected vertex $u$, we search for semantically compatible features
within a local foreground neighbourhood $\mathcal{N}_i(u)$.
The similarity between the baked vertex descriptor $\mathbf{f}_u$ and an image
feature is measured by cosine similarity:
\begin{equation}
s_{u,p}
=
\left\langle
\overline{\mathbf{f}}_u,
\overline{F}_i(p)
\right\rangle,
\qquad
p\in\mathcal{N}_i(u),
\end{equation}
where the overline denotes $L_2$ normalisation.
We retain only vertices whose best local match exceeds a similarity threshold
$\eta$, thereby removing unreliable correspondences.
Rather than selecting a hard nearest neighbour, we use a soft weighting
$\alpha_{u,p}\propto\exp(s_{u,p}/\tau)$ over the valid local matches.
The resulting correspondence loss is:
\begin{equation}
\label{eq:dino_loss}
L_{\mathrm{RSC}}
=
\frac{1}{|\widehat{\mathcal{U}}_i|}
\sum_{u\in\widehat{\mathcal{U}}_i}
\sum_{p\in\mathcal{N}_i(u)\cap M_i}
\alpha_{u,p}\left(1-s_{u,p}\right),
\end{equation}
where $\widehat{\mathcal{U}}_i$ is the set of reliable visible vertices.
This local, confidence-filtered matching provides semantic guidance for pose
optimization while reducing the influence of unstable correspondences.

\section{PAW4D Dataset}
\label{sec:dataset}
\label{sec:dataset}
A substantial challenge for 4D animal reconstruction is the lack of datasets with ground-truth 3D annotations, particularly in world coordinates. To address this gap, we introduce PAW4D, a synthetic yet realistic benchmark with known 3D geometry and camera parameters. We build on DeformingThings4D~\citep{li20214dcomplete}, which provides a large collection of animated 4D meshes. We retain only realistic animal sequences, excluding atypical species (e.g., dragons), physically implausible motions (e.g., swimming on land), and sequences shorter than one second. Since the original textures are unavailable, we retexture the selected meshes using EmbodyGen~\citep{wang2025embodiedgengenerative3dworld}, which synthesizes plausible textures from a mesh and a text prompt. Each sequence is rendered in an outdoor environment with a flat ground plane and an HDRI background under three camera configurations: \textit{follow}, where the camera tracks the animal; \textit{fixed}, where the camera remains stationary after the first frame; and \textit{orbit}, where the camera moves around the scene. We additionally apply small Gaussian perturbations to the \textit{follow} and \textit{fixed} cameras, while randomly sampling camera azimuth, elevation, initial position, and distance. PAW4D comprises 115 videos, including 51 dog, 47 fox, 13 puma, and 4 bear sequences. Each video contains RGB frames, ground-truth masks, known cameras, and per-frame ground-truth meshes.


\section{Experiments}
\label{sec:Experiments}
\noindent\textbf{Datasets.}
For evaluation, we report results on PFERD~\citep{Li2024PFERD} and our PAW4D. For PFERD, we randomly choose 26 videos from the 130 videos, covering all five horses in the dataset. For PAW4D, we use 109 of the 115 sequences for evaluation and reserve the remaining six sequences for ablation studies.

\noindent\textbf{Baselines.} 
We compare our method with four recent state-of-the-art (SOTA) methods. AniMer~\citep{lyu2025animer} and GenZoo~\citep{niewiadomski2025ICCVgenzoo} are multi-species model-based approaches that reconstruct 3D animal from images. We additionally compare with 3D Fauna~\citep{Li20243Dfauna}, a model-free animal reconstruction method, and its video-based extension 4D Fauna~\citep{zhao2025webscaleanimal4dfauna}, which incorporates temporal information for 4D animal reconstruction.

\noindent\textbf{Evaluation Metrics.}
For 2D appearance, we report Intersection over Union (IoU), Peak Signal-to-Noise Ratio (PSNR), and Learned Perceptual Image Patch Similarity (LPIPS). For 3D geometry, we use volumetric Intersection over Union (IoU3D) and Chamfer Distance (CD). On PFERD, which provides hSMAL ground-truth meshes and joints, we additionally report Procrustes-Aligned Mean Per-Vertex Position Error (PA-MPVPE) and Procrustes-Aligned Mean Per-Joint Position Error (PA-MPJPE) to evaluate 3D reconstruction accuracy. Finally, we evaluate world-space motion using trajectory Root Mean Square Error (RMSE).

\noindent\textbf{Implementation Details.}
We use an Adam optimizer for 300 epochs with early stopping. All our experiments run on a machine equipped with an H100 GPU. 


\newlength{\baselineH}
\newsavebox{\baselinebox}

\begin{figure}[t]
    \centering

    \settoheight{\baselineH}{%
        \includegraphics{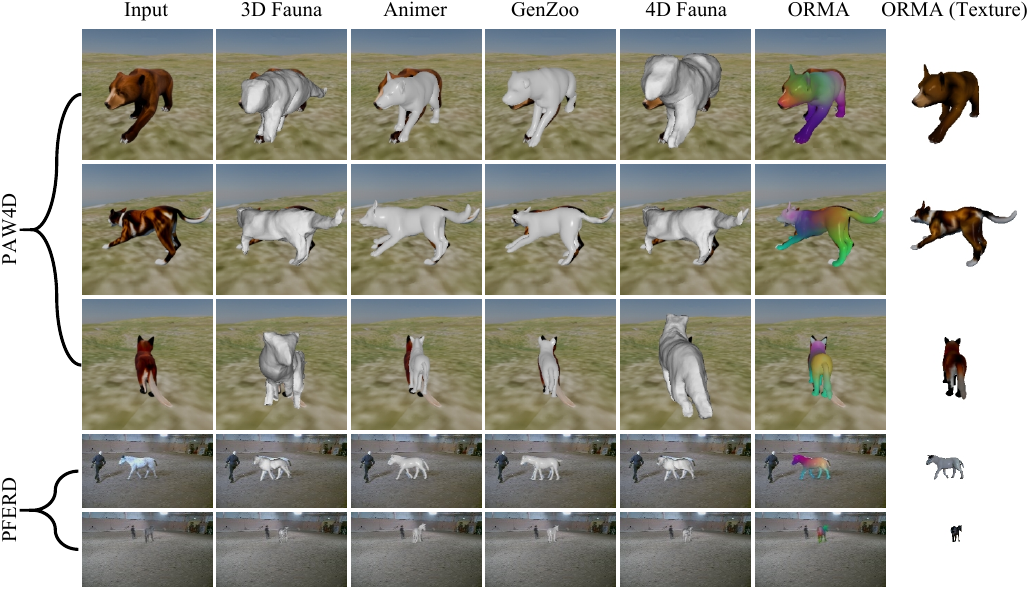}%
    }

    \sbox{\baselinebox}{%
        \includegraphics[
            width=0.8\linewidth,
            trim=0 0 0 {.04\baselineH},
            clip
        ]{figures/baseline_results.pdf}%
    }

    \newcommand*{\basecoltitle}[2]{%
        \put(#1,\ht\baselinebox+1pt){%
            \makebox(0,0)[b]{%
                \scriptsize
                \begin{tabular}{@{}c@{}}#2\end{tabular}%
            }%
        }%
    }

    \vspace{2em}

    \begin{Overpic}{\usebox{\baselinebox}}
        \basecoltitle{14.3}{Input} 
        \basecoltitle{27.2}{3D Fauna}
        \basecoltitle{40.5}{AniMer}
        \basecoltitle{53.5}{GenZoo}
        \basecoltitle{66.5}{4D Fauna}
        \basecoltitle{79.5}{ORMA}
        \basecoltitle{93.5}{ORMA (Texture)}
    \end{Overpic}

    \caption{
        Qualitative comparisons on the PAW4D and PFERD datasets.
        We compare our method with AniMer~\citep{lyu2025animer},
        GenZoo~\citep{niewiadomski2025ICCVgenzoo},
        3D Fauna~\citep{Li20243Dfauna}, and
        4D Fauna~\citep{zhao2025webscaleanimal4dfauna}.
        AniMer, GenZoo, and 3D Fauna operate on individual images,
        whereas 4D Fauna and ORMA take videos as input and exploit
        temporal information across frames.
    }
    \label{fig:baseline_results}
    \vspace{-0.3cm}
\end{figure}


\newlength{\pferdH}
\newsavebox{\pferdbox}

\begin{figure}[t]
    \centering

    \settoheight{\pferdH}{%
        \includegraphics{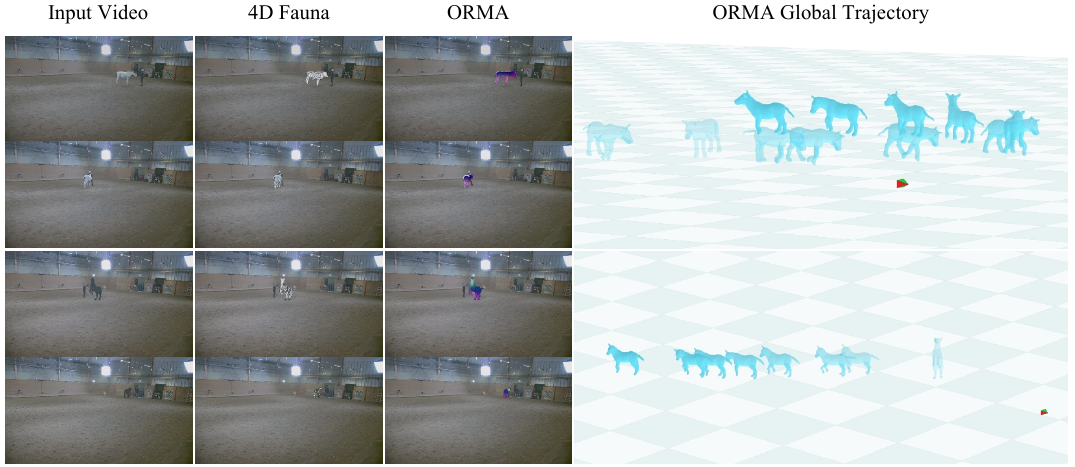}%
    }

    \sbox{\pferdbox}{%
        \includegraphics[
            width=\linewidth,
            trim=0 0 0 {.06\pferdH},
            clip
        ]{figures/pferd_result5.pdf}%
    }

    \newcommand*{\pferdcoltitle}[2]{%
        \put(#1,\ht\pferdbox+1pt){%
            \makebox(0,0)[b]{%
                \small
                \begin{tabular}{@{}c@{}}#2\end{tabular}%
            }%
        }%
    }

    \vspace{2em}

    \begin{Overpic}{\usebox{\pferdbox}}
        \pferdcoltitle{9.3}{Input Video}
        \pferdcoltitle{26.8}{4D Fauna}
        \pferdcoltitle{44.3}{ORMA}
        \pferdcoltitle{75.6}{ORMA Global Trajectory}
    \end{Overpic}

    \caption{
        We compare ORMA with 4D-Fauna on two videos from PFERD.
        ORMA better aligns the reconstructed animal with the observations across
        frames, while additionally recovering its motion in a shared world
        coordinate frame. The rightmost column visualizes the resulting global
        trajectory.
    }
    \label{fig:pferd_results}
    \vspace{-0.3cm}
\end{figure}


\newlength{\ormaH}
\newsavebox{\ormabox}
\begin{figure}[t]
    \centering
    \settoheight{\ormaH}{\includegraphics{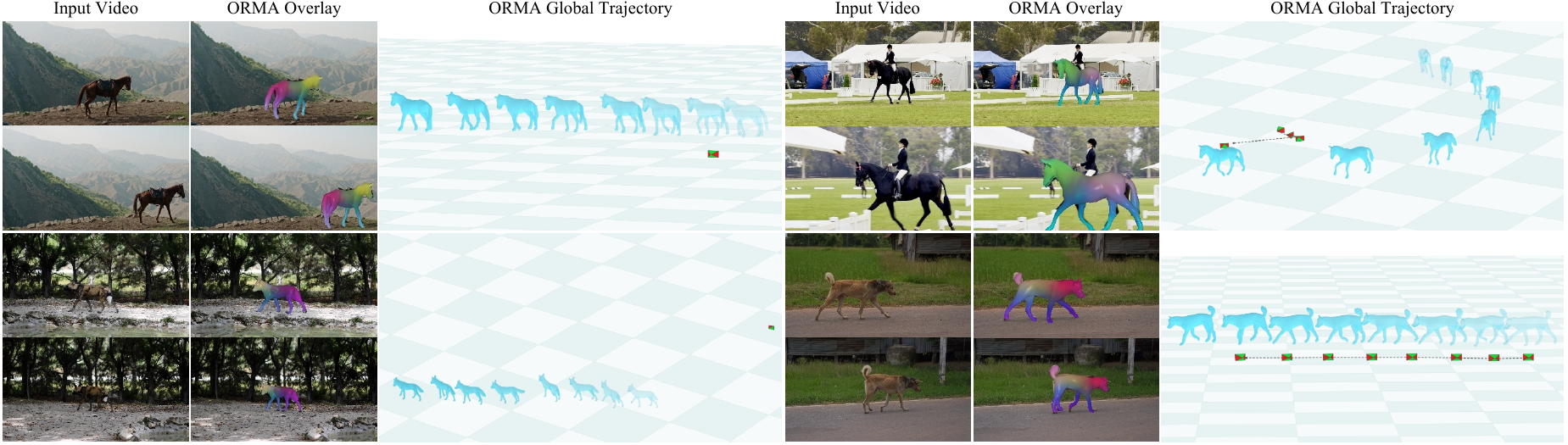}}
    \sbox{\ormabox}{\includegraphics[width=\linewidth, trim=0 0 0 {.075\ormaH}, clip]{figures/demo_horizon.pdf}}
    \newcommand*{\coltitle}[2]{\put(#1,\ht\ormabox+1pt){\makebox(0,0)[b]{\scriptsize\begin{tabular}{@{}c@{}}#2\end{tabular}}}}
    \vspace{2em}
    \begin{Overpic}{\usebox{\ormabox}}
        \coltitle{6.3}{Input Video}
        \coltitle{18.0}{ORMA Overlay}
        \coltitle{36.9}{ORMA Global Trajectory}
        \coltitle{56.3}{Input Video}
        \coltitle{68.0}{ORMA Overlay}
        \coltitle{86.9}{ORMA Global Trajectory}
    \end{Overpic}
    \caption{Qualitative evaluation of ORMA under challenging in-the-wild scenarios, including large variations in viewpoint, scale, and camera motion. The overlay and global trajectory visualisations show that ORMA robustly reconstructs articulated animal motion and world-space displacement under unconstrained capture conditions.}
    \label{fig:demo_results}
    \vspace{-0.3cm}
\end{figure}

\begin{table*}[t]
    \centering
    \caption{
        \textbf{Comparison with state-of-the-art animal reconstruction methods on PFERD~\citep{Li2024PFERD}.}
        $\uparrow$ indicates higher is better, and $\downarrow$ indicates lower is better.
    }
    \label{tab:PFERD}

    \setlength{\tabcolsep}{5.2pt}
    \renewcommand{\arraystretch}{1.10}
    \scriptsize

    \begin{adjustbox}{max width=\textwidth}
    \begin{tabular}{@{}ccccccccc@{}}
        \toprule
        \multirow{2}{*}{Method}
        & \multicolumn{3}{c}{2D Appearance}
        & \multicolumn{2}{c}{3D Geometry}
        & Trajectory
        & \multicolumn{2}{c}{3D Pose} \\
        \cmidrule(lr){2-4}
        \cmidrule(lr){5-6}
        \cmidrule(lr){7-7}
        \cmidrule(lr){8-9}
        \noalign{\vskip -2pt}

        & IoU $\uparrow$
        & PSNR $\uparrow$
        & LPIPS $\downarrow$
        & IoU3D $\uparrow$
        & Chamfer $\downarrow$
        & RMSE $\downarrow$
        & PA-MPVPE $\downarrow$
        & PA-MPJPE $\downarrow$ \\
        \midrule

        3D Fauna~\citep{Li20243Dfauna}
        & 0.500 & 24.314 & 0.028
        & 0.130 & 0.714
        & --
        & 467.9 & 450.2 \\

        AniMer~\citep{lyu2025animer}
        & \textbf{0.676} & 25.370 & 0.020
        & 0.204 & 0.407
        & --
        & 193.6 & 251.4 \\

        GenZoo~\citep{niewiadomski2025ICCVgenzoo}
        & 0.562 & 25.845 & 0.023
        & 0.184 & 0.437
        & --
        & 198.5 & \underline{246.8} \\

        4D Fauna~\citep{zhao2025webscaleanimal4dfauna}
        & 0.497 & 24.315 & 0.028
        & 0.130 & 0.715
        & --
        & 476.7 & 458.1 \\

        \midrule

        Ours
        & 0.601 & \underline{27.474} & \underline{0.018}
        & \underline{0.348} & \underline{0.149}
        & \underline{1.587}
        & \underline{181.8} & 250.5 \\

        Ours (w/ GT Cam)
        & \underline{0.655} & \textbf{27.785} & \textbf{0.017}
        & \textbf{0.367} & \textbf{0.130}
        & \textbf{1.073}
        & \textbf{176.1} & \textbf{246.2} \\

        \bottomrule
    \end{tabular}
    \end{adjustbox}
    \vspace{-0.3cm}
\end{table*}

\begin{table*}[t]
    \centering
    \caption{
        \textbf{Comparison with state-of-the-art animal reconstruction methods on our synthetic dataset PAW4D.}
        $\uparrow$ indicates higher is better, and $\downarrow$ indicates lower is better.
    }
    \label{tab:PAW4D}
    \setlength{\tabcolsep}{5.2pt}
    \renewcommand{\arraystretch}{1.10}
    \scriptsize
    \begin{adjustbox}{max width=\textwidth}
    \begin{tabular}{@{}ccccccc@{}}
        \toprule
        \multirow{2}{*}{Method}
        & \multicolumn{3}{c}{2D Appearance}
        & \multicolumn{2}{c}{3D Geometry}
        & Trajectory \\
        \cmidrule(lr){2-4}
        \cmidrule(lr){5-6}
        \cmidrule(lr){7-7}
        \noalign{\vskip -2pt}
        & IoU $\uparrow$
        & PSNR $\uparrow$
        & LPIPS $\downarrow$
        & IoU3D $\uparrow$
        & Chamfer $\downarrow$
        & RMSE $\downarrow$ \\
        \midrule

        3D Fauna~\citep{Li20243Dfauna}
        & 0.495 & 14.540 & 0.255
        & 0.190 & \underline{0.040}
        & -- \\

        AniMer~\citep{lyu2025animer}
        & 0.668 & 14.690 & 0.140
        & 0.196 & 0.046
        & -- \\

        GenZoo~\citep{niewiadomski2025ICCVgenzoo}
        & 0.704 & 14.470 & 0.139
        & 0.212 & 0.049
        & -- \\

        4D Fauna~\citep{zhao2025webscaleanimal4dfauna}
        & 0.368 & 12.990 & 0.316
        & 0.200 & 0.042
        & -- \\

        \midrule

        Ours
        & \underline{0.714} & \underline{22.260} & \underline{0.137}
        & \underline{0.215} & 0.045
        & \underline{0.131} \\

        Ours (w/ GT Cam)
        & \textbf{0.729} & \textbf{22.300} & \textbf{0.134}
        & \textbf{0.422} & \textbf{0.006}
        & \textbf{0.046} \\

        \bottomrule
    \end{tabular}
    \end{adjustbox}
    
\end{table*}

\subsection{Comparison}
\label{sec:comparison}

\noindent\textbf{Comparison without GT camera.}
We report quantitative results in Tables~\ref{tab:PFERD} and~\ref{tab:PAW4D}. Using MegaSaM-estimated cameras, ORMA performs strongly on both datasets, substantially improving 3D reconstruction on PFERD and achieving the best 2D appearance metrics and highest IoU3D among the compared methods on PAW4D. This suggests improved geometric accuracy without sacrificing image alignment. Trajectory RMSE is reported only for ORMA, as the baselines do not recover camera-to-world transformations or global animal trajectories.

\noindent\textbf{Qualitative comparison.}
Figs.~\ref{fig:baseline_results} and~\ref{fig:pferd_results} show qualitative comparisons on PAW4D and PFERD. Model-based methods such as AniMer and GenZoo recover plausible poses but are limited by their parametric shape spaces, whereas 3D Fauna and 4D Fauna allow more flexible geometry but can suffer from pose or alignment errors. ORMA combines instance-specific generative geometry with structured SMAL{+} articulation and robust semantic correspondences, improving shape fidelity while maintaining accurate pose and image alignment. Additional in-the-wild results in Fig.~\ref{fig:demo_results} further demonstrate generalization across diverse animal appearances, poses, and camera motions.
\begin{table}[t]
\centering
\caption{\textbf{Ablation study of the individual components of ORMA.}}
\label{tab:orma_ablation}

\fontsize{6.5}{7.2}\selectfont
\setlength{\tabcolsep}{3.0pt}
\renewcommand{\arraystretch}{1.05}

\begin{adjustbox}{max width=\columnwidth}
\begin{tabular}{ccccccc}
\toprule
\multirow{2}{*}{Variant}
& \multicolumn{3}{c}{2D Appearance}
& \multicolumn{2}{c}{3D Geometry}
& \multicolumn{1}{c}{Trajectory} \\
\cmidrule(lr){2-4}
\cmidrule(lr){5-6}
\cmidrule(lr){7-7}
& IoU$\uparrow$
& PSNR$\uparrow$
& LPIPS$\downarrow$
& IoU3D$\uparrow$
& Chamfer$\downarrow$
& RMSE$\downarrow$ \\
\midrule

w/o Hunyuan3D
& 0.756 & 22.45 & 0.1503
& 0.409 & 0.005 & 0.043 \\

w/o DINO
& 0.755 & 22.63 & 0.1482
& 0.411 & 0.004 & 0.041 \\

w/o SMAL+
& 0.744 & 22.62 & 0.1500
& 0.408 & 0.005 & 0.040 \\

w/o Depth
& 0.706 & 22.51 & 0.1522
& 0.313 & 0.016 & 0.148 \\

w/o Temporal Smoothing
& 0.755 & 22.63 & 0.1482
& 0.412 & 0.005 & 0.040 \\

w/o Temporal \& Depth
& 0.706 & 22.52 & 0.1522
& 0.312 & 0.016 & 0.148 \\

\midrule

Full
& \textbf{0.761} & \textbf{22.64} & \textbf{0.1481}
& \textbf{0.414} & \textbf{0.005} & \textbf{0.039} \\

\bottomrule
\end{tabular}
\end{adjustbox}

\end{table}

\begin{figure*}[!t] 
    \centering
    \includegraphics[width=0.9\linewidth]{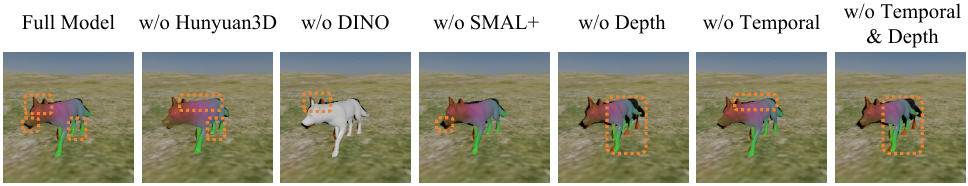}
    \caption{Ablation studies. Hunyuan3D, DINO,  SMAL+ and other optimization losses each lead to improved performance, as discussed in Sec.~\ref{sec:ablation}. 
    \label{fig:ablation}}
    \vspace{-0.3cm}
\end{figure*}

\newlength{\ablationH}
\newsavebox{\ablationbox}








\subsection{Ablation}
\label{sec:ablation}
Table~\ref{tab:orma_ablation} and Fig.~\ref{fig:ablation} evaluate the contribution of each component of ORMA.
Depth supervision has the largest impact: removing it substantially degrades both
3D reconstruction and trajectory accuracy, reducing IoU3D from 0.414 to 0.313
and increasing trajectory RMSE from 0.039 to 0.148.
Removing Hunyuan3D or replacing SMAL{+} with the original SMAL consistently reduces reconstruction quality, highlighting the benefits of instance-specific geometry and the richer SMAL+ shape space. DINO correspondences provide additional improvements in image alignment and
3D consistency, while temporal smoothing yields consistent gains in
trajectory accuracy. Overall, the full model achieves the strongest performance across all metrics,
indicating that the proposed components provide complementary supervision for
appearance, geometry, and global motion reconstruction.

\section{Conclusions}
\label{sec:conclusions}
We presented \method{}, an optimization-based framework for monocular 4D reconstruction of articulated animals. \method{} combines instance-specific geometry from generative 3D priors with the structured articulation of SMAL+, and jointly leverages geometric and semantic supervision to recover temporally coherent animal motion in a consistent world coordinate frame. Experiments demonstrate improved reconstruction accuracy and robust global motion reconstruction across diverse quadruped species. We further introduce PAW4D, a synthetic multi-species benchmark with ground-truth geometry and camera motion for quantitative evaluation of 4D animal reconstruction.


\noindent\textbf{Limitations.}
\method{} assumes that the target animal can be registered to the SMAL+ topology, limiting its applicability to quadrupeds with compatible morphology; substantially different body plans, such as birds or aquatic animals, are not supported. The method also relies on accurate camera estimation and does not explicitly model severe occlusions, making reference frames with substantial self-occlusion difficult to initialize. Finally, the SMAL+ representation relies on Linear Blend Skinning, which cannot fully capture strong pose- or motion-dependent non-rigid deformations.
Extending the framework to more diverse animal topologies and dynamic deformations is an important direction for future work.



\clearpage

\subsubsection*{Acknowledgments}
This work was supported by the European Research Council (ERC) Advanced Grant SIMULACRON and by the GNI Project "AI4Twinning".

\bibliography{iclr2027_conference}

@String(IJCV  = {IJCV})

@String(CVPR  = {CVPR})

@String(ICCV  = {ICCV})

@String(ECCV  = {ECCV})

@incollection{SMPL:2015,
  title={SMPL: A skinned multi-person linear model},
  author={Loper, Matthew and Mahmood, Naureen and Romero, Javier and Pons-Moll, Gerard and Black, Michael J},
  booktitle={Seminal Graphics Papers: Pushing the Boundaries, Volume 2},
  pages={851--866},
  year={2023}
}

@inproceedings{ruegg2023bite,
  title={Bite: Beyond priors for improved three-d dog pose estimation},
  author={R{\"u}egg, Nadine and Tripathi, Shashank and Schindler, Konrad and Black, Michael J and Zuffi, Silvia},
  booktitle={2023 IEEE/CVF Conference on Computer Vision and Pattern Recognition (CVPR)},
  pages={8867--8876},
  year={2023},
  organization={IEEE}
}

@inproceedings{zhong2026_4danimal,
  title={4d-animal: Freely reconstructing animatable 3d animals from videos},
  author={Zhong, Shanshan and Peng, Jiawei and Zheng, Zehan and Huang, Zhongzhan and Ma, Wufei and Zhang, Guofeng and Liu, Qihao and Yuille, Alan and Chen, Jieneng},
  booktitle={2026 IEEE/CVF Winter Conference on Applications of Computer Vision (WACV)},
  pages={602--612},
  year={2026},
  organization={IEEE}
}

@inproceedings{cmrKanazawa18,
  title={Learning category-specific mesh reconstruction from image collections},
  author={Kanazawa, Angjoo and Tulsiani, Shubham and Efros, Alexei A and Malik, Jitendra},
  booktitle={European Conference on Computer Vision},
  pages={386--402},
  year={2018},
  organization={Springer}
}

@article{Yao2022lassie,
  title={Lassie: Learning articulated shapes from sparse image ensemble via 3d part discovery},
  author={Yao, Chun-Han and Hung, Wei-Chih and Li, Yuanzhen and Rubinstein, Michael and Yang, Ming-Hsuan and Jampani, Varun},
  journal={Advances in Neural Information Processing Systems},
  volume={35},
  pages={15296--15308},
  year={2022}
}

@inproceedings{Wu2023magicpony,
  title={Magicpony: Learning articulated 3d animals in the wild},
  author={Wu, Shangzhe and Li, Ruining and Jakab, Tomas and Rupprecht, Christian and Vedaldi, Andrea},
  booktitle={2023 IEEE/CVF Conference on Computer Vision and Pattern Recognition (CVPR)},
  pages={8792--8802},
  year={2023},
  organization={IEEE}
}

@inproceedings{Li20243Dfauna,
  title={Learning the 3d fauna of the web},
  author={Li, Zizhang and Litvak, Dor and Li, Ruining and Zhang, Yunzhi and Jakab, Tomas and Rupprecht, Christian and Wu, Shangzhe and Vedaldi, Andrea and Wu, Jiajun},
  booktitle={2024 IEEE/CVF Conference on Computer Vision and Pattern Recognition (CVPR)},
  pages={9752--9762},
  year={2024},
  organization={IEEE}
}

@article{yang2021viser,
  title={Viser: Video-specific surface embeddings for articulated 3d shape reconstruction},
  author={Yang, Gengshan and Sun, Deqing and Jampani, Varun and Vlasic, Daniel and Cole, Forrester and Liu, Ce and Ramanan, Deva},
  journal={Advances in Neural Information Processing Systems},
  volume={34},
  pages={19326--19338},
  year={2021}
}

@article{yang2021lasr,
  title={Lasr: Learning articulated shape reconstruction from a monocular video},
  author={Yang, Gengshan and Sun, Deqing and Jampani, Varun and Vlasic, Daniel and Cole, Forrester and Chang, Huiwen and Ramanan, Deva and Freeman, William T and Liu, Ce},
  journal={arXiv preprint arXiv:2105.02976},
  year={2021}
}

@inproceedings{BANMO,
  title={Banmo: Building animatable 3d neural models from many casual videos},
  author={Yang, Gengshan and Vo, Minh and Neverova, Natalia and Ramanan, Deva and Vedaldi, Andrea and Joo, Hanbyul},
  booktitle={2022 IEEE/CVF Conference on Computer Vision and Pattern Recognition (CVPR)},
  pages={2853--2863},
  year={2022},
  organization={IEEE}
}

@inproceedings{yang2023ppr,
  title={Ppr: Physically plausible reconstruction from monocular videos},
  author={Yang, Gengshan and Yang, Shuo and Zhang, John Z and Manchester, Zachary and Ramanan, Deva},
  booktitle={2023 IEEE/CVF International Conference on Computer Vision (ICCV)},
  pages={3891--3901},
  year={2023},
  organization={IEEE}
}

@inproceedings{Zuffi_2019_ICCV,
  title={Three-d safari: Learning to estimate zebra pose, shape, and texture from images “in the wild”},
  author={Zuffi, Silvia and Kanazawa, Angjoo and Berger-Wolf, Tanya and Black, Michael},
  booktitle={2019 IEEE/CVF International Conference on Computer Vision (ICCV)},
  pages={5358--5367},
  year={2019},
  organization={IEEE}
}

@inproceedings{Zuffi:CVPR:2017,
  title={3D menagerie: Modeling the 3D shape and pose of animals},
  author={Zuffi, Silvia and Kanazawa, Angjoo and Jacobs, David W and Black, Michael J},
  booktitle={2017 IEEE conference on computer vision and pattern recognition (CVPR)},
  pages={5524--5532},
  year={2017},
  organization={IEEE}
}

@article{Li2024PFERD,
  title={The poses for equine research dataset (pferd)},
  author={Li, Ci and Mellbin, Ylva and Krogager, Johanna and Polikovsky, Senya and Holmberg, Martin and Ghorbani, Nima and Black, Michael J and Kjellstr{\"o}m, Hedvig and Zuffi, Silvia and Hernlund, Elin},
  journal={Scientific Data},
  volume={11},
  number={1},
  pages={497},
  year={2024},
  publisher={Nature Publishing Group UK London}
}

@inproceedings{zuffi2024awol,
  title={Awol: Analysis without synthesis using language},
  author={Zuffi, Silvia and Black, Michael J},
  booktitle={European Conference on Computer Vision},
  pages={1--19},
  year={2024},
  organization={Springer}
}

@inproceedings{Zuffi_2024_CVPR,
  title={Varen: Very accurate and realistic equine network},
  author={Zuffi, Silvia and Mellbin, Ylva and Li, Ci and Hoeschle, Markus and Kjellstr{\"o}m, Hedvig and Polikovsky, Senya and Hernlund, Elin and Black, Michael J},
  booktitle={Proceedings of the IEEE/CVF Conference on Computer Vision and Pattern Recognition},
  pages={5374--5383},
  year={2024}
}

@inproceedings{badger_2020_eccv,
  title={3D bird reconstruction: a dataset, model, and shape recovery from a single view},
  author={Badger, Marc and Wang, Yufu and Modh, Adarsh and Perkes, Ammon and Kolotouros, Nikos and Pfrommer, Bernd G and Schmidt, Marc F and Daniilidis, Kostas},
  booktitle={European conference on computer vision},
  pages={1--17},
  year={2020},
  organization={Springer}
}

@inproceedings{Wang_2021_CVPR,
  title={Birds of a feather: Capturing avian shape models from images},
  author={Wang, Yufu and Kolotouros, Nikos and Daniilidis, Kostas and Badger, Marc},
  booktitle={2021 IEEE/CVF Conference on Computer Vision and Pattern Recognition (CVPR)},
  pages={14734--14744},
  year={2021},
  organization={IEEE}
}

@inproceedings{biggs2020left,
  title={Who left the dogs out? 3d animal reconstruction with expectation maximization in the loop},
  author={Biggs, Benjamin and Boyne, Oliver and Charles, James and Fitzgibbon, Andrew and Cipolla, Roberto},
  booktitle={European Conference on Computer Vision},
  pages={195--211},
  year={2020},
  organization={Springer}
}

@article{li2021hsmaldetailedhorseshape,
  title={hsmal: Detailed horse shape and pose reconstruction for motion pattern recognition},
  author={Li, Ci and Ghorbani, Nima and Broom{\'e}, Sofia and Rashid, Maheen and Black, Michael J and Hernlund, Elin and Kjellstr{\"o}m, Hedvig and Zuffi, Silvia},
  journal={arXiv preprint arXiv:2106.10102},
  year={2021}
}

@article{baieri2025modelbasedmetric3dshape,
  title={Model-based metric 3d shape and motion reconstruction of wild bottlenose dolphins in drone-shot videos},
  author={Baieri, Daniele and Cicciarella, Riccardo and Kr{\"u}tzen, Michael and Rodol{\`a}, Emanuele and Zuffi, Silvia},
  journal={International Journal of Computer Vision},
  volume={134},
  number={6},
  pages={293},
  year={2026},
  publisher={Springer}
}

@inproceedings{biggs_2019,
  title={Creatures great and smal: Recovering the shape and motion of animals from video},
  author={Biggs, Benjamin and Roddick, Thomas and Fitzgibbon, Andrew and Cipolla, Roberto},
  booktitle={Asian Conference on Computer Vision},
  pages={3--19},
  year={2018},
  organization={Springer}
}

@inproceedings{AnimalAvatars2024,
  title={Animal avatars: Reconstructing animatable 3d animals from casual videos},
  author={Sabathier, Remy and Mitra, Niloy J and Novotny, David},
  booktitle={European Conference on Computer Vision},
  pages={270--287},
  year={2024},
  organization={Springer}
}

@inproceedings{SMPL-X:2019,
  title={Expressive body capture: 3d hands, face, and body from a single image},
  author={Pavlakos, Georgios and Choutas, Vasileios and Ghorbani, Nima and Bolkart, Timo and Osman, Ahmed AA and Tzionas, Dimitrios and Black, Michael J},
  booktitle={Proceedings of the IEEE/CVF conference on computer vision and pattern recognition},
  pages={10975--10985},
  year={2019}
}

@article{BEDLAM2,
  title={BEDLAM2. 0: Synthetic humans and cameras in motion},
  author={Tesch, Joachim and Becherini, Giorgio and Achar, Prerana and Yiannakidis, Anastasios and Kocabas, Muhammed and Patel, Priyanka and Black, Michael},
  journal={Advances in Neural Information Processing Systems},
  volume={38},
  year={2026}
}

@article{zhao2025hunyuan3d20,
  title={Hunyuan3d 2.0: Scaling diffusion models for high resolution textured 3d assets generation},
  author={Zhao, Zibo and Lai, Zeqiang and Lin, Qingxiang and Zhao, Yunfei and Liu, Haolin and Yang, Shuhui and Feng, Yifei and Yang, Mingxin and Zhang, Sheng and Yang, Xianghui and others},
  journal={arXiv preprint arXiv:2501.12202},
  year={2025}
}

@inproceedings{lyu2025animer,
  title={Animer: Animal pose and shape estimation using family aware transformer},
  author={Lyu, Jin and Zhu, Tianyi and Gu, Yi and Lin, Li and Cheng, Pujin and Liu, Yebin and Tang, Xiaoying and An, Liang},
  booktitle={2025 IEEE/CVF Conference on Computer Vision and Pattern Recognition (CVPR)},
  pages={17486--17496},
  year={2025},
  organization={IEEE}
}

@inproceedings{niewiadomski2025ICCVgenzoo,
  title={Generative zoo},
  author={Niewiadomski, Tomasz and Yiannakidis, Anastasios and Cuevas-Velasquez, Hanz and Sanyal, Soubhik and Black, Michael J and Zuffi, Silvia and Kulits, Peter},
  booktitle={2025 IEEE/CVF International Conference on Computer Vision (ICCV)},
  pages={8492--8502},
  year={2025},
  organization={IEEE}
}

@inproceedings{li2025megasam,
  title={Megasam: Accurate, fast, and robust structure and motion from casual dynamic videos},
  author={Li, Zhengqi and Tucker, Richard and Cole, Forrester and Wang, Qianqian and Jin, Linyi and Ye, Vickie and Kanazawa, Angjoo and Holynski, Aleksander and Snavely, Noah},
  booktitle={2025 IEEE/CVF Conference on Computer Vision and Pattern Recognition (CVPR)},
  pages={10486--10496},
  year={2025},
  organization={IEEE}
}

@article{zhao2025webscaleanimal4dfauna,
  title={Web-scale collection of video data for 4d animal reconstruction},
  author={Zhao, Brian Nlong and Wu, Jiajun and Wu, Shangzhe},
  journal={Advances in Neural Information Processing Systems},
  volume={38},
  year={2026}
}

@article{lyu2025animerunifiedposeshape+,
  title={Animer+: Unified pose and shape estimation across mammalia and aves via family-aware transformer},
  author={An, Liang and Lyu, Jin and Lin, Li and Cheng, Pujin and Liu, Yebin and Tang, Xiaoying},
  journal={IEEE Transactions on Pattern Analysis and Machine Intelligence},
  year={2025},
  publisher={IEEE}
}

@inproceedings{kirillov2023segment,
  title={Segment anything},
  author={Kirillov, Alexander and Mintun, Eric and Ravi, Nikhila and Mao, Hanzi and Rolland, Chloe and Gustafson, Laura and Xiao, Tete and Whitehead, Spencer and Berg, Alexander C and Lo, Wan-Yen and others},
  booktitle={2023 IEEE/CVF international conference on computer vision (ICCV)},
  pages={3992--4003},
  year={2023},
  organization={IEEE}
}

@inproceedings{sorkine2007rigid,
  title={As-rigid-as-possible surface modeling},
  author={Sorkine, Olga and Alexa, Marc},
  booktitle={Symposium on Geometry processing},
  volume={4},
  pages={109--116},
  year={2007}
}

@inproceedings{wimmer2025anyup,
  title={Anyup: Universal feature upsampling},
  author={Wimmer, Thomas and Truong, Prune and Rakotosaona, Marie-Julie and Oechsle, Michael and Tombari, Federico and Schiele, Bernt and Lenssen, Jan Eric},
  booktitle={International Conference on Learning Representations},
  volume={2026},
  pages={140700--140720},
  year={2026}
}

@inproceedings{rueegg2022barc,
  title={Barc: Learning to regress 3d dog shape from images by exploiting breed information},
  author={Rueegg, Nadine and Zuffi, Silvia and Schindler, Konrad and Black, Michael J},
  booktitle={2022 IEEE/CVF Conference on Computer Vision and Pattern Recognition (CVPR)},
  pages={3866--3874},
  year={2022},
  organization={IEEE}
}

@inproceedings{li20214dcomplete,
  title={4dcomplete: Non-rigid motion estimation beyond the observable surface},
  author={Li, Yang and Takehara, Hikari and Taketomi, Takafumi and Zheng, Bo and Nie{\ss}ner, Matthias},
  booktitle={2021 IEEE/CVF International Conference on Computer Vision (ICCV)},
  pages={12686--12696},
  year={2021},
  organization={IEEE}
}

@article{wang2025embodiedgengenerative3dworld,
  title={Embodiedgen: Towards a generative 3d world engine for embodied intelligence},
  author={Wang, Xinjie and Liu, Liu and Cao, Yu and Wu, Ruiqi and Qin, Wenkang and Wang, Dehui and Sui, Wei and Su, Zhizhong},
  journal={arXiv preprint arXiv:2506.10600},
  year={2025}
}

@inproceedings{yang2023rac,
  title={Reconstructing animatable categories from videos},
  author={Yang, Gengshan and Wang, Chaoyang and Reddy, N Dinesh and Ramanan, Deva},
  booktitle={2023 IEEE/CVF Conference on Computer Vision and Pattern Recognition (CVPR)},
  pages={16995--17005},
  year={2023},
  organization={IEEE}
}

@inproceedings{LiaoPADR,
  title={Pad3r: Pose-aware dynamic 3d reconstruction from casual videos},
  author={Liao, Ting-Hsuan and Liu, Haowen and Xu, Yiran and Ge, Songwei and Yang, Gengshan and Huang, Jia-Bin},
  booktitle={Proceedings of the SIGGRAPH Asia 2025 Conference Papers},
  pages={1--11},
  year={2025}
}

@inproceedings{kulkarni2019csm,
  title={Canonical surface mapping via geometric cycle consistency},
  author={Kulkarni, Nilesh and Gupta, Abhinav and Tulsiani, Shubham},
  booktitle={Proceedings of the ieee/cvf international conference on computer vision},
  pages={2202--2211},
  year={2019}
}

@article{lyu20264dequine,
  title={4DEquine: Disentangling Motion and Appearance for 4D Equine Reconstruction from Monocular Video},
  author={Lyu, Jin and An, Liang and Cheng, Pujin and Liu, Yebin and Tang, Xiaoying},
  journal={arXiv preprint arXiv:2603.10125},
  year={2026}
}

@article{yang2022apt,
  title={Apt-36k: A large-scale benchmark for animal pose estimation and tracking},
  author={Yang, Yuxiang and Yang, Junjie and Xu, Yufei and Zhang, Jing and Lan, Long and Tao, Dacheng},
  journal={Advances in Neural Information Processing Systems},
  volume={35},
  pages={17301--17313},
  year={2022}
}

@inproceedings{sinha2023common,
  title={Common pets in 3d: Dynamic new-view synthesis of real-life deformable categories},
  author={Sinha, Samarth and Shapovalov, Roman and Reizenstein, Jeremy and Rocco, Ignacio and Neverova, Natalia and Vedaldi, Andrea and Novotny, David},
  booktitle={2023 IEEE/CVF Conference on Computer Vision and Pattern Recognition (CVPR)},
  pages={4881--4891},
  year={2023},
  organization={IEEE}
}

@inproceedings{mathis2021pretraining,
  title={Pretraining boosts out-of-domain robustness for pose estimation},
  author={Mathis, Alexander and Biasi, Thomas and Schneider, Steffen and Yuksekgonul, Mert and Rogers, Byron and Bethge, Matthias and Mathis, Mackenzie W},
  booktitle={Proceedings of the IEEE/CVF winter conference on applications of computer vision},
  pages={1859--1868},
  year={2021}
}

@article{li2019atrw,
  title={ATRW: A benchmark for amur tiger re-identification in the wild},
  author={Li, Shuyuan and Li, Jianguo and Tang, Hanlin and Qian, Rui and Lin, Weiyao},
  journal={arXiv preprint arXiv:1906.05586},
  year={2019}
}

@article{labuguen2021macaquepose,
  title={MacaquePose: a novel “in the wild” macaque monkey pose dataset for markerless motion capture},
  author={Labuguen, Rollyn and Matsumoto, Jumpei and Negrete, Salvador Blanco and Nishimaru, Hiroshi and Nishijo, Hisao and Takada, Masahiko and Go, Yasuhiro and Inoue, Ken-ichi and Shibata, Tomohiro},
  journal={Frontiers in behavioral neuroscience},
  volume={14},
  pages={581154},
  year={2021},
  publisher={Frontiers}
}

@article{wah2011caltech,
  title={The caltech-ucsd birds-200-2011 dataset},
  author={Wah, Catherine and Branson, Steve and Welinder, Peter and Perona, Pietro and Belongie, Serge},
  year={2011}
}

@article{yu2021ap,
  title={Ap-10k: A benchmark for animal pose estimation in the wild},
  author={Yu, Hang and Xu, Yufei and Zhang, Jing and Zhao, Wei and Guan, Ziyu and Tao, Dacheng},
  journal={arXiv preprint arXiv:2108.12617},
  year={2021}
}

@article{banik2021novel,
  title={A novel dataset for keypoint detection of quadruped animals from images},
  author={Banik, Prianka and Li, Lin and Dong, Xishuang},
  journal={arXiv preprint arXiv:2108.13958},
  year={2021}
}

@inproceedings{ng2022animal,
  title={Animal kingdom: A large and diverse dataset for animal behavior understanding},
  author={Ng, Xun Long and Ong, Kian Eng and Zheng, Qichen and Ni, Yun and Yeo, Si Yong and Liu, Jun},
  booktitle={2022 IEEE/CVF Conference on Computer Vision and Pattern Recognition (CVPR)},
  pages={19001--19012},
  year={2022},
  organization={IEEE}
}

@inproceedings{kearney2020rgbd,
  title={Rgbd-dog: Predicting canine pose from rgbd sensors},
  author={Kearney, Sinead and Li, Wenbin and Parsons, Martin and Kim, Kwang In and Cosker, Darren},
  booktitle={2020 IEEE/CVF Conference on Computer Vision and Pattern Recognition (CVPR)},
  pages={8333--8342},
  year={2020},
  organization={IEEE}
}

@inproceedings{joska2021acinoset,
  title={Acinoset: a 3d pose estimation dataset and baseline models for cheetahs in the wild},
  author={Joska, Daniel and Clark, Liam and Muramatsu, Naoya and Jericevich, Ricardo and Nicolls, Fred and Mathis, Alexander and Mathis, Mackenzie W and Patel, Amir},
  booktitle={2021 IEEE international conference on robotics and automation (ICRA)},
  pages={13901--13908},
  year={2021},
  organization={IEEE}
}

@inproceedings{xu2023animal3d,
  title={Animal3d: A comprehensive dataset of 3d animal pose and shape},
  author={Xu, Jiacong and Zhang, Yi and Peng, Jiawei and Ma, Wufei and Jesslen, Artur and Ji, Pengliang and Hu, Qixin and Zhang, Jiehua and Liu, Qihao and Wang, Jiahao and others},
  booktitle={2023 IEEE/CVF International Conference on Computer Vision (ICCV)},
  pages={9065--9075},
  year={2023},
  organization={IEEE}
}

@inproceedings{choi2026everydog,
  title     = {Every Dog Has Its Day, Probably: A Balanced Synthetic Benchmark and
               Probabilistic Modeling for 3D Dog Pose Estimation},
  author    = {Choi, Joo Young and Lee, Wonkwang and Seon, Ju-hyeong and Kim, Gunhee},
  booktitle = {Proceedings of the European Conference on Computer Vision (ECCV)},
  year      = {2026},
}

@article{aamir2026wilddepth,
  title={WildDepth: A Multimodal Dataset for 3D Wildlife Perception and Depth Estimation},
  author={Aamir, Muhammad and Muramatsu, Naoya and Shin, Sangyun and Wijers, Matthew and Zhong, Jia-Xing and Hou, Xinyu and Patel, Amir and Loveridge, Andrew and Markham, Andrew},
  journal={arXiv preprint arXiv:2603.16816},
  year={2026}
}

@inproceedings{cao2019cross,
  title={Cross-domain adaptation for animal pose estimation},
  author={Cao, Jinkun and Tang, Hongyang and Fang, Hao-Shu and Shen, Xiaoyong and Lu, Cewu and Tai, Yu-Wing},
  booktitle={Proceedings of the IEEE/CVF international conference on computer vision},
  pages={9498--9507},
  year={2019}
}

@inproceedings{shooter2024digidogs,
  title={Digidogs: Single-view 3d pose estimation of dogs using synthetic training data},
  author={Shooter, Moira and Malleson, Charles and Hilton, Adrian},
  booktitle={2024 IEEE/CVF Winter Conference on Applications of Computer Vision Workshops (WACVW)},
  pages={92--101},
  year={2024},
  organization={IEEE}
}

@article{hu2026sam,
  title={SAM 3D Animal: Promptable Animal 3D Reconstruction from Images in the Wild},
  author={Hu, Xuyi and Lyu, Jin and Liu, Jiuming and Liu, Yebin and Zuffi, Silvia and An, Liang and Goetz, Stefan},
  journal={arXiv preprint arXiv:2605.07604},
  year={2026}
}

@inproceedings{cho2026wildani4d,
  title={WildAni4D: Towards 4D Animal Mesh Reconstruction},
  author={Cho, Gyeongsu and Hu, Hezhen and Soon, Donghyeon and Kang, Changwoo and Joo, Kyungdon},
  booktitle={Proceedings of the IEEE/CVF Conference on Computer Vision and Pattern Recognition},
  pages={160--169},
  year={2026}
}

@article{wang2025dogmo,
  title={DogMo: A Large-Scale Multi-View RGB-D Dataset for 4D Canine Motion Recovery},
  author={Wang, Zan and Chen, Siyu and Mo, Luya and Gao, Xinfeng and Shen, Yuxin and Ding, Lebin and Liang, Wei},
  journal={arXiv preprint arXiv:2510.24117},
  year={2025}
}

@article{peng2026interpet4d,
  title={InterPet4D: A Multimodal 4D Human-Pet Interaction Dataset for Pet Motion Generation},
  author={Peng, Yichen and Song, Jyun-Ting and Liao, Chen-Chieh and Kitani, Kris and Koike, Hideki and Wu, Erwin},
  journal={arXiv preprint arXiv:2607.10287},
  year={2026}
}

@misc{cao2026redirect4dbench,
  title  = {Redirect4D-Bench: A Scalable Benchmark for Camera Redirection of Monocular Dynamic Videos with Pseudo-4D Ground Truth},
  author = {Wei Cao and Hao Zhang and Jiapeng Tang and Yulun Wu and Yingying Li and Ning Yu and Shenlong Wang and Yaoyao Liu},
  year   = {2026},
}

@article{yu2026prima,
  title={PRIMA: Boosting Animal Mesh Recovery with Biological Priors and Test-Time Adaptation},
  author={Yu, Xiaohang and Wang, Ti and Mathis, Mackenzie Weygandt},
  journal={arXiv preprint arXiv:2606.02366},
  year={2026}
}

@article{simeoni2025dinov3,
  title={Dinov3},
  author={Sim{\'e}oni, Oriane and Vo, Huy V and Seitzer, Maximilian and Baldassarre, Federico and Oquab, Maxime and Jose, Cijo and Khalidov, Vasil and Szafraniec, Marc and Yi, Seungeun and Ramamonjisoa, Micha{\"e}l and others},
  journal={arXiv preprint arXiv:2508.10104},
  year={2025}
}

@inproceedings{zhao2026kirin,
  title={Kirin: Animal Motion Generation from In-the-Wild Video},
  author={Zhao, Brian Nlong and Pan, Zhuoyang and Rehg, James M and Wu, Jiajun and Wu, Shangzhe},
  booktitle={European Conference on Computer Vision},
  pages={1--19},
  year={2026},
  organization={Springer}
}

@InProceedings{sun2024ponymation,
  title     = {{Ponymation}: Learning Articulated 3D Animal Motions from Unlabeled Online Videos},
  author    = {Keqiang Sun and Dor Litvak and Yunzhi Zhang and Hongsheng Li and Jiajun Wu and Shangzhe Wu},
  booktitle = ECCV,
  year      = {2024}
}

@Article{wu2023dove,
  title     = {{DOVE}: Learning Deformable 3D Objects by Watching Videos},
  author    = {Shangzhe Wu and Tomas Jakab and Christian Rupprecht and Andrea Vedaldi},
  journal   = {IJCV},
  year      = {2023}
}

@article{an2023three,
  title={Three-dimensional surface motion capture of multiple freely moving pigs using MAMMAL},
  author={An, Liang and Ren, Jilong and Yu, Tao and Hai, Tang and Jia, Yichang and Liu, Yebin},
  journal={Nature Communications},
  volume={14},
  number={1},
  pages={7727},
  year={2023},
  publisher={Nature Publishing Group UK London}
}

@inproceedings{ye2023decoupling,
  title={Decoupling human and camera motion from videos in the wild},
  author={Ye, Vickie and Pavlakos, Georgios and Malik, Jitendra and Kanazawa, Angjoo},
  booktitle={2023 IEEE/CVF Conference on Computer Vision and Pattern Recognition (CVPR)},
  pages={21222--21232},
  year={2023},
  organization={IEEE}
}

@article{sinha2022common,
  title={Common Pets in 3D: Dynamic New-View Synthesis of Real-Life Deformable Categories},
  author={Sinha, Samarth and Shapovalov, Roman and Reizenstein, Jeremy and Rocco, Ignacio and Neverova, Natalia and Vedaldi, Andrea and Novotny, David},
  journal={CVPR},
  year={2023}
}
\bibliographystyle{iclr2027_conference}

\clearpage
\appendix
\section{Appendix}
\subsection{More Details about PAW4D}
A major challenge for evaluating animal reconstruction methods is the lack of datasets with reliable 3D ground-truth annotations in world coordinates. Existing datasets such as COP3D~\citep{sinha2022common} contain videos of real animals but do not provide ground-truth geometry or camera calibration, making it difficult to quantitatively verify whether reconstructed motion is correct in global space or to measure errors such as scale drift and temporally inconsistent poses.

To enable controlled evaluation, we generate a synthetic benchmark with full geometric supervision. Our dataset is built from DeformingThings4D~\citep{li20214dcomplete}, which contains animated meshes represented as sequences of per-vertex deformation offsets applied to a canonical base mesh. Applying these offsets sequentially reconstructs the mesh animation describing the animal motion. The original dataset contains 1,972 deformation sequences across 26 categories.
Following the filtering procedure described in the main paper, we discard sequences that are unsuitable for terrestrial outdoor scenes, including aquatic animals, physically implausible motions, or sequences shorter than one second. After filtering, 121 candidate animations remain.

\begin{figure}
    \centering
    \includegraphics[width=\linewidth]{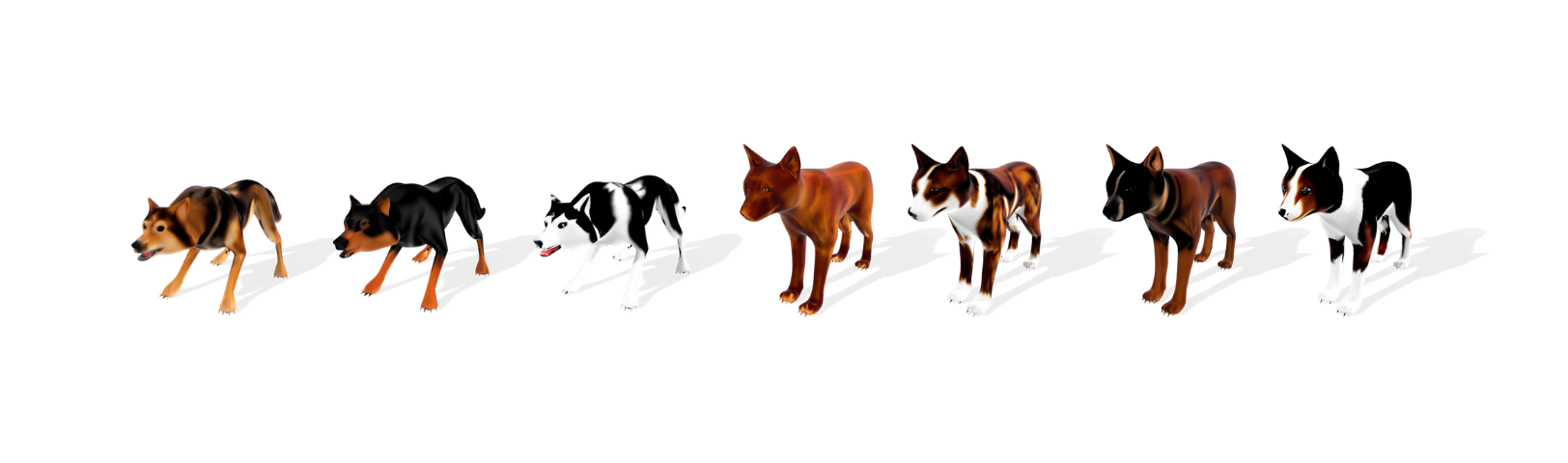}
    \caption{Examples of generated textures.}
    \label{fig:textures}
\end{figure}

Because the original textures were not released, we generate new textures using EmbodyGen~\citep{wang2025embodiedgengenerative3dworld}, which synthesizes plausible textures conditioned on a mesh and a text prompt. Multiple texture variations are generated for each animal category to increase appearance diversity (Fig.~\ref{fig:textures}).

All sequences are rendered in a synthetic outdoor scene consisting of grassy terrain and HDRI sky illumination (note that, to ensure that the terrain provides realistic cues for depth and camera estimation, we use a 3D terrain asset rendered under the scene camera). For each animation we render three camera configurations: follow (camera tracks the animal), fixed (camera remains static), and orbit (camera rotates around the scene). Camera azimuth, elevation, distance, and initial position are randomly sampled, and the follow and fixed configurations additionally include small Gaussian camera jitter.

For every frame, we export RGB images, segmentation masks, depth maps, and the ground-truth mesh in world coordinates, together with the full camera calibration parameters (intrinsics and camera-to-world transformations).


\begin{table}[H]
\centering
\small
\setlength{\tabcolsep}{6pt}
\begin{tabular}{lccccc}
\toprule
& \multicolumn{5}{c}{Species} \\
\cmidrule(lr){2-6}
Asset & Dog & Fox & Puma & Bear & Total \\
\midrule
Animations & 51 & 47 & 13 & 4 & \textbf{115} \\
Textures   & 13 & 2  & 3  & 1 & \textbf{19} \\
\bottomrule
\end{tabular}
\caption{Statistics of the animal assets used in our synthetic dataset PAW4D, including the number of animation sequences and generated texture variants.}
\label{tab:paw4d_stats}
\end{table}

\subsection{Loss Functions}

\noindent\textbf{2D Geometric Loss.}
The 2D geometric loss encourages the projected template to align with the
2D observation in frame $\image{i}$ and consists of three terms:
\begin{equation}
\label{eq:app}
L_{\mathrm{app}}
=
w_{\mathrm{dice}}L_{\mathrm{dice}}
+w_{\mathrm{per}}L_{\mathrm{per}}
+w_{\mathrm{bound}}L_{\mathrm{bound}}.
\end{equation}

Let $M_i^{\mathrm{rend}}$ denote the rendered soft silhouette of $S_{H3D}(\pose_i,\mathbf{R}_{g,i},\trasl_i)$ under camera $C_i$, and let $M_i$ denote the corresponding SAM mask.
The first term is a Dice loss that aligns the projected silhouette with the
observed foreground mask:
\begin{equation}
L_{\mathrm{dice}}
=
1-
\frac{
2\langle M_i^{\mathrm{rend}},M_i\rangle
}{
\|M_i^{\mathrm{rend}}\|_1+\|M_i\|_1+\epsilon
}.
\end{equation}

While the Dice loss encourages overall foreground overlap, it is less sensitive
to contour discrepancies. We therefore introduce two complementary boundary
terms. Let $\mathcal{E}(\cdot)$ denote a differentiable edge-magnitude map
computed using Sobel gradients.

The first encourages the rendered and observed silhouettes to have similar
overall boundary magnitude:
\begin{equation}
L_{\mathrm{per}}
=
\rho\!\left(
\frac{\|\mathcal{E}(M_i^{\mathrm{rend}})\|_1}
     {\|\mathcal{E}(M_i)\|_1+\epsilon}
-1
\right).
\end{equation}

The second directly penalizes local misalignment between their boundaries:
\begin{equation}
L_{\mathrm{bound}}
=
\rho\!\left(
\mathcal{E}(M_i^{\mathrm{rend}})-\mathcal{E}(M_i)
\right),
\end{equation}
where $\rho$ denotes the Smooth-$L_1$ penalty.

\begin{figure*}[t] 
    \centering
    \includegraphics[width=0.6\textheight]{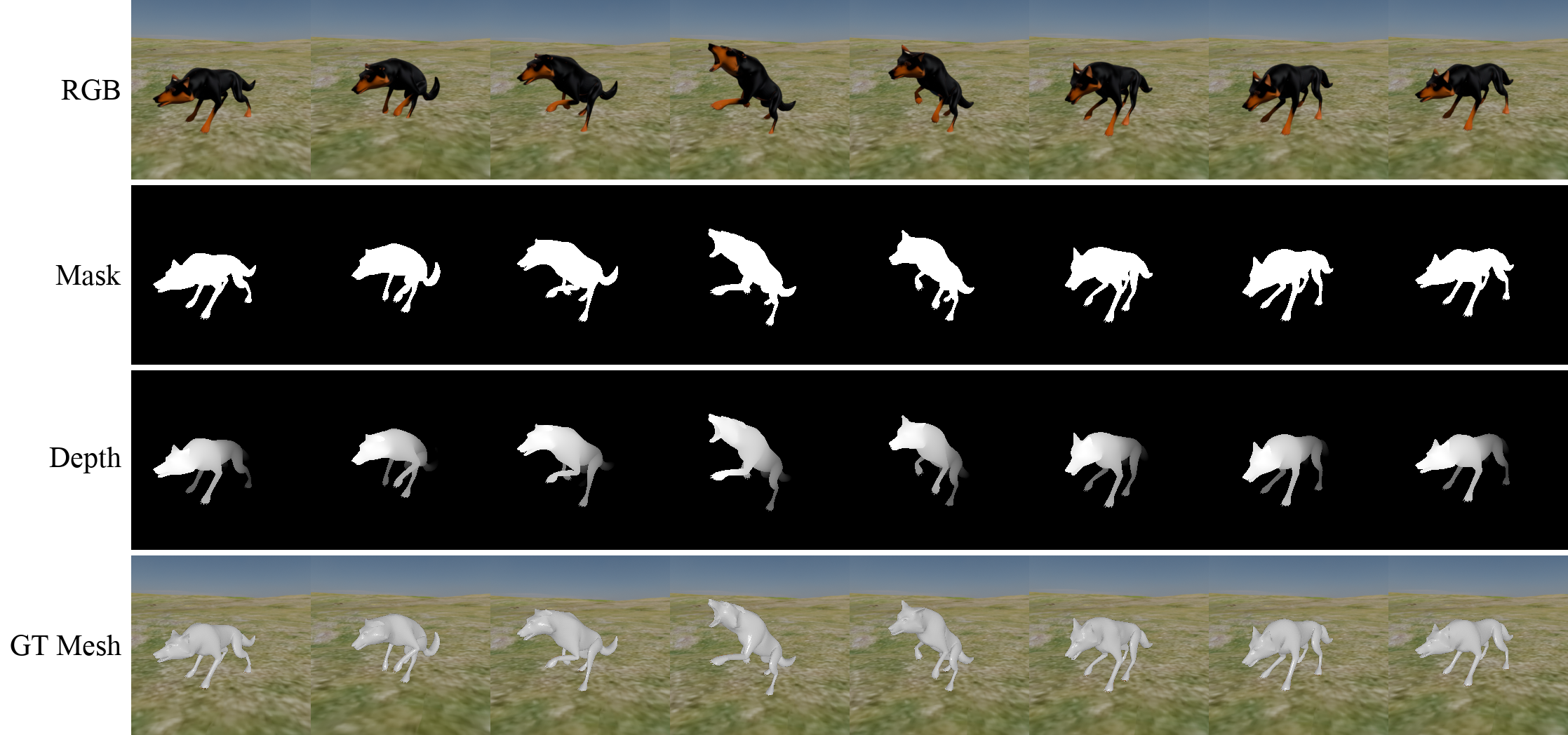}
    \caption{PAW4D Dataset. \label{fig:PAW4D}}
\end{figure*}

\noindent\textbf{Depth Loss.}
MegaSaM provides dense depth estimates that complement the image-space
objectives with explicit 3D geometric supervision. We use two complementary
depth constraints:
\begin{equation}
\label{eq:depth}
L_{\mathrm{depth}}
=
w_{\mathrm{root}}L_{\mathrm{root}}
+
w_{\mathrm{surf}}L_{\mathrm{surf}}.
\end{equation}
The surface-depth term $L_{\mathrm{surf}}$ aligns the rendered animal surface
with the calibrated MegaSaM depth over valid foreground pixels, while the
root-depth term $L_{\mathrm{root}}$ constrains the camera-space position of
the animal using the foreground depth. Together, these terms provide
complementary supervision for local surface alignment and global 3D placement.

\noindent\textbf{Regularisation Losses.}
AniMer{+} provides a useful pose prior that prevents the optimization from
drifting toward implausible configurations:
\begin{equation}
\label{eq:prior}
    L_{\text{pose}}
    = MSE(\pose, \pose_{animer{+}}),
\end{equation}

Temporal consistency is imposed on second-order motion:
\begin{equation}
\label{eq:temporal}
L_{\mathrm{temp}}
=
w_{\mathrm{trans}}L_{\mathrm{trans}}
+
w_{\mathrm{rot}}L_{\mathrm{rot}}
+
w_{\mathrm{joint}}L_{\mathrm{joint}},
\end{equation}
where the three terms penalize acceleration in world-space translation,
global rotation, and root-relative joint motion, respectively.

\subsection{Robust Semantic Correspondence}

Our robust semantic correspondences are derived from the textured Hunyuan3D reconstruction, whose texture provides stable visual cues for matching. We render a set of canonical views of the reconstructed mesh and extract dense DINO features from each view. Because these features capture fine-scale appearance details, such as fur patterns and local markings, they provide reliable cues for animal correspondence.

\begin{figure}[t]
    \centering
    \includegraphics[width=0.4\linewidth]{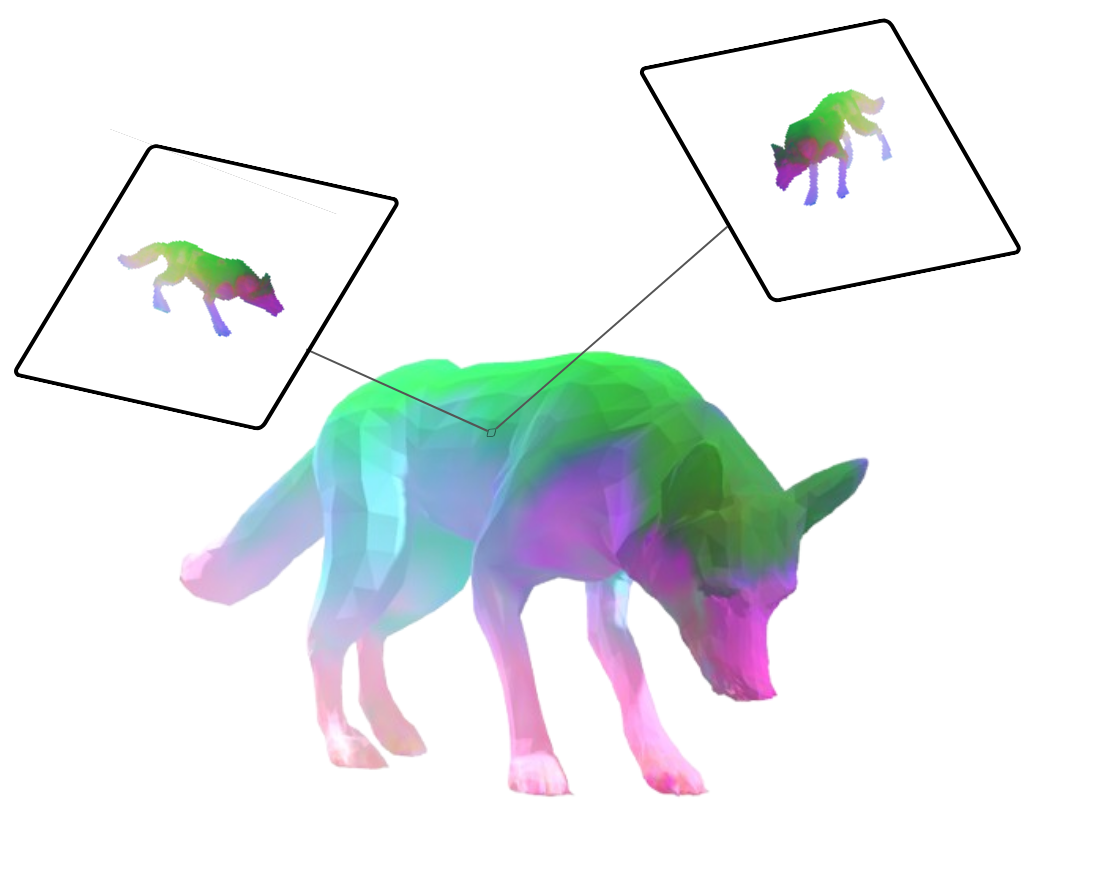}
    \caption{We aggregate the DINO features from multiple views on the 3D shape surface. Such features are then used to establish a correspondence between the 3D shape and the video frames.}
    \label{fig:DINO}
\end{figure}

These features are then projected back onto the reconstructed surface to obtain per-vertex descriptors, which are subsequently transferred to the SMAL+ model. In Fig.~\ref{fig:DINO} we depict a visualization of feature aggregation on the animal's 3D surface. This produces a dense semantic representation on the SMAL surface, guiding the optimization toward semantically meaningful alignments. Additional details are provided in Sec.~\ref{sec:RSC}. 

\begin{figure}[t]
    \centering
    \includegraphics[width=\linewidth]{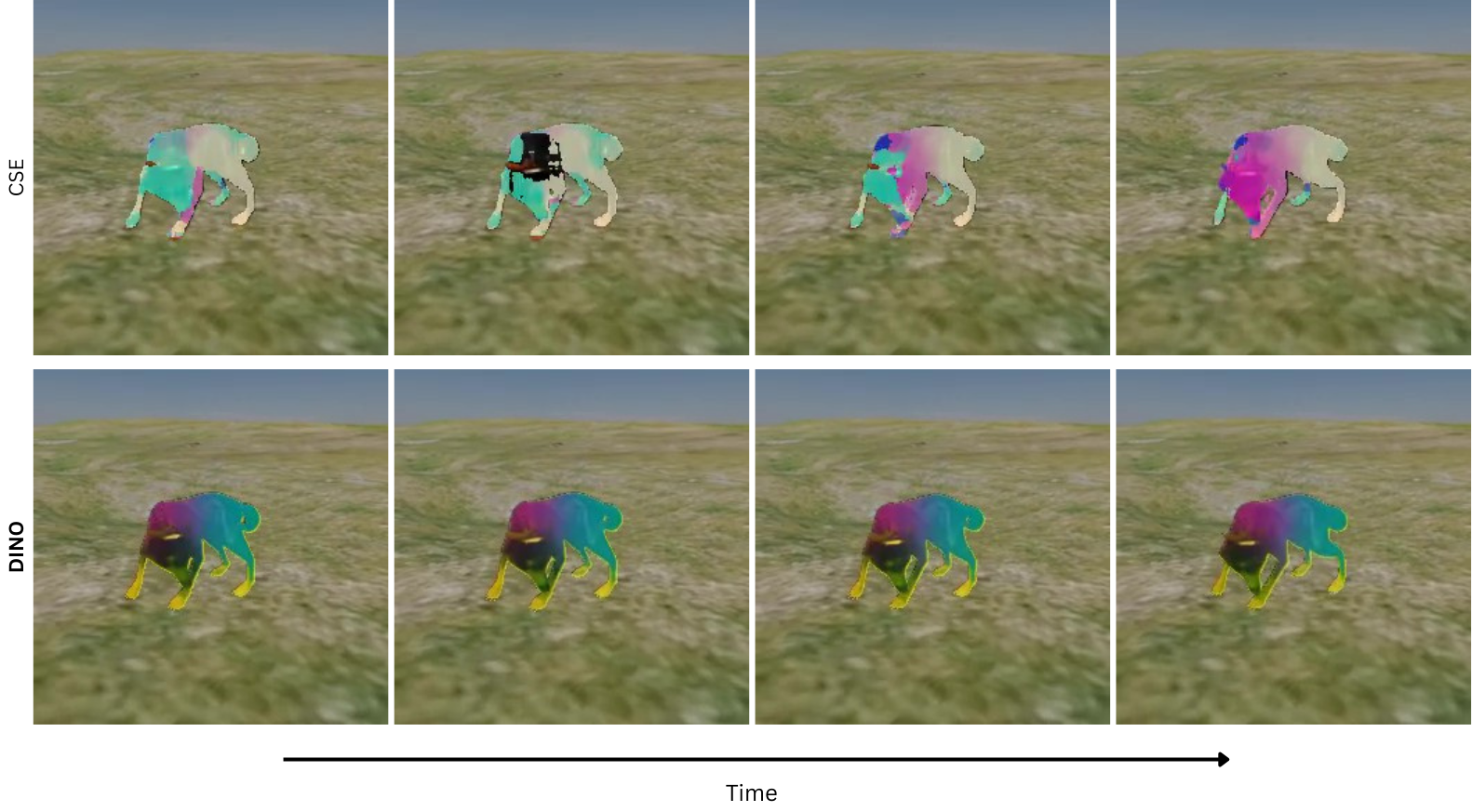}
    \caption{Comparison between CSE features (top) and our DINO-based features (bottom). Note how the CSE features vary across visually similar frames.}
    \label{fig:CSE}
\end{figure}

Compared to CSE-based correspondences~\citep{kulkarni2019csm}, this representation is substantially more stable over time. In practice, CSE predictions often vary across frames even under limited motion, whereas DINO-based features remain much more consistent, especially in articulated or visually distinctive regions. As shown in Fig.~\ref{fig:CSE}, this improved temporal stability leads to more reliable correspondences and reduces optimization drift.

\end{document}